\documentclass[sigconf]{acmart}
\renewcommand\footnotetextcopyrightpermission[1]{} 

\AtBeginDocument{%
	\providecommand\BibTeX{{%
			\normalfont B\kern-0.5em{\scshape i\kern-0.25em b}\kern-0.8em\TeX}}}

\setcopyright{acmcopyright}
\copyrightyear{2020}
\acmYear{2020}
\acmDOI{ DOI NUMBER } 

\acmConference[NLPIR 2020]{NLPIR 2020: 4th International Conference on Natural Language Processing and Information Retrieval
}{June 26--28, 2020}{Seoul, Korea}
\acmBooktitle{NLPIR 2020: 4th International Conference on Natural Language Processing and Information Retrieval, June 26--28, 2020, Seoul, Korea}
\acmPrice{15.00} 
\acmISBN{978-1-4503-7760-7} 
\usepackage{graphicx}
\usepackage{comment}
\usepackage{soul}
\newcommand{\ngram}{\textit{n}\mbox{-}gram}
\newcommand{\qgram}{\textit{q}\mbox{-}gram}
\usepackage{algorithm,algpseudocode}

\begin{document}
	
	\title{Novel Keyword Extraction \& Language Detection Approaches}
	
	\author{Malgorzata Pikies}
	\email{malgorzata@cloudflare.com}
	\affiliation{%
		\institution{Cloudflare}
		\city{London}
		\country{United Kingdom}
	}
	
	\author{Andronicus Riyono}
	\email{andronicus@cloudflare.com}
	\affiliation{%
		\institution{Cloudflare}
		\city{Singapore}
		\country{Singapore}
	}

	\author{Junade Ali}
	\email{junade@cloudflare.com}
	\affiliation{%
		\institution{Cloudflare}
		\city{London}
		\country{United Kingdom}
	}

	\renewcommand{\shortauthors}{Pikies, Riyono and Ali}
	
	\begin{abstract}
		Fuzzy string matching and language classification are important tools in Natural Language Processing pipelines, this paper provides advances in both areas.
		We propose a fast novel approach to string tokenisation for fuzzy language matching and experimentally demonstrate an $83.6\%$ decrease in processing time with an estimated improvement in recall of $3.1\%$ at the cost of a $2.6\%$ decrease in precision. This approach is able to work even where keywords are subdivided into multiple words, without needing to scan character-to-character.
		So far there has been little work considering using metadata to enhance language classification algorithms. We provide observational data and find the \texttt{Accept-Language} header is $14\%$ more likely to match the classification than the country language associated to the IP Address.
	\end{abstract}

	

	\pagestyle{plain} 
	\maketitle
	
	\section{Introduction}
	
	Feature extraction is an important part of Natural Language Processing pipelines, whether extracting important keywords \cite{8820497} or language detection \cite{castro2017smoothed}.
	
During analysis; text data can be divided into sentences, words, or characters, which can then later be treated individually or can be gathered into groups of \ngram s \cite{Kondrak:2005:NGS:2178997.2179011}. Fuzzy string matching compares tokenised text to keywords using string similarity algorithms. For example; the Edit Distance (also known as the Levenshtein distance) \cite{levenshtein1966bcc} measures the number of elementary changes necessary needed to transform one string into another, using the following elementary edit operations:
	\begin{itemize}
		\item a change operation, if $X \neq \emptyset$ and $Y \neq \emptyset$;
		\item a delete operation, if $Y = \emptyset$;
		\item an insert operation, if $X = \emptyset$.
	\end{itemize}
	 Other string similarity algorithms like Cosine similarity \cite{snlp} and Dice \cite{Kondrak:2005:NGS:2178997.2179011}, divide strings into sets of letters known as \ngram s. Given these algorithms are not sensitive to the order of characters and \ngram s, the size of the \ngram can dramatically alter the accuracy of the algorithm. Equation \ref{eq:cos} shows a formula for calculating the Cosine similarity between strings X and Y. Strings are divided into \ngram s, where each unique \ngram~ is a separate dimension in a multi-dimensional vector space. The two vectors made of strings X and Y are then used to calculate the cosine of the angle between them:
	\begin{equation} 	
	s(X,Y) = \frac{\vec{U}(X) \cdot \vec{V}(Y)}{|\vec{U}(X)||\vec{V}(Y) |} = cos \theta.
	\label{eq:cos}
	\end{equation}
	
	\ngram \ based approaches can also be used for language detection, but are notably unreliable on short corpuses of text \cite{castro2017smoothed}. Whilst internet standards and web browsers have sought to standardise content language headers \cite{rfc3282}, there has been little study on the accuracy of these fields.

	Our prior work \cite{8820497} provided a more detailed definition of various string similarity algorithms and provided empirical analysis of their performance for fuzzy string matching. We extend upon this work here by developing a novel tokenisation approach for fuzzy string matching. We propose a novel hybrid approach to approximate keyword search. 
	Our focus is on the impact on accuracy and computation speed while using a greedy approach for sub-string selection prior to its tokenisation. We further explore using internet user metadata to improve language classification.
	
	In Section \ref{sec:lit} we describe research papers related to language detection and keyword search including \ngram s. In Section \ref{sec:greedy} we describe our hybrid method to a keyword search text classification. Section \ref{sec:lang} explores how metadata like web browser headers and IP Addresses can be used to improve the accuracy of language classification algorithms. We summarise our research and present conclusions in Section \ref{sec:conclu}.
	
	\section{Literature Review}
	\label{sec:lit}
	
	There are many algorithms designed to find exact matches of strings, such as Knuth-Morris-Pratt \cite{Knuthf74fastpattern} or the Boyer Moore algorithm \cite{Boyer:1977:FSS:359842.359859}. \cite{Li2007VGRAMIP} developed a novel technique to generate variable-length grams (VGRAMs) and showed that VGRAM tokenisation improved performance of three chosen algorithms. Additionally, \cite{Litwin:2007:FNS:1325851.1325878} describe a novel approach for  \ngram ~- based string search in the 'write once read many' context. Their algorithm uses \ngram~ signatures together with an algorithm similar to the Boyer Moore algorithm thus their technique is also focused on exact string matching. 
	
	In our approach, we instead wish to perform fuzzy string matching rather than exact matching. \cite{fuzzyTokenMatching} proposed a fuzzy-token similarity metric, which is a combination of token and character based similarities. The algorithm looks for a maximum sum of weights between pairs of tokens in two strings from a weighted bigraph. They also proposed an efficient method based on tokens' signatures called Fast-Join. 
	
	\cite{ExtendingGrams} proposed using a wildcard symbol (it can represent any character from the alphabet) in \qgram s. They proposed two algorithms, \textit{BasicEQ} and \textit{OptEQ}, that use a concept of string hierarchy, combinatorial analysis, and semi-lattice for selectivity estimation. In \cite{Lee:2009:ASS:1516360.1516455} authors proposed two algorithms, \textit{The MOst Frequent Minimal Base String Method} (MOF) and \textit{Lower Bound Estimation} (LBS), to perform an estimation of selectivity of approximate substring queries based an extended \ngram \ table with wildcards.
	
	\cite{Cavnar1994NgrambasedTC} showed that the \ngram \ based frequency method is both inexpensive and effective in documents classification. They split the text into \ngram s of sizes from one to five (letters) and counted their occurrences using a hash table. 
	
	
	 
	 \ngram s can also be used in language classification. \cite{castro2017smoothed} considers using smoothed \ngram \ based models for language identification of Twitter messages - the authors compared a smoothed n-gram language model with a TF-IDF weighting scheme alongside comparing various classifiers (Naive-Bayes, Logistic Regression, SVM, and LLR classifiers). The authors conclude that: "This study validates the fact that when it comes to dealing with very short texts we need to conduct deep investigations based on this domain."
	 
	 \cite{wu2019enriching} incorporates entity level information into a pre-trained language model, but to the best of our knowledge there is no such work incorporating metadata into language classification models. \cite{graham2014world} found that "found that there are significant challenges to accurately determining the language of tweets in an automated manner" but notes challenges of using purely geolocation data for language classification.
	 
	 \cite{rfc3282} provides that web browsers may pass language preferences to websites using the \texttt{Accept-Language} header; whilst this has been implemented in modern web browsers, there has been no empirical study of the accuracy of such language headers.
	 
	 \cite{stiller2010ambiguity} experimented with using IP Address information and user interface language to predict the language used in user input forms on a small sample of $510$ logs; the authors note that these features alone are not strong indicators for determining query language and more robust dimensions are needed. \cite{leveling2010dcu} found a correlation between country language, interface language and input language; but to the authors surprise, found only $24\%$ of queries were in a language associated with the user's country (obtained from their IP Address), the work did not consider other browser headers and the logs were from a pan-European online library (it is not understood if the context affected the language input of users). We could not identify prior work seeking to use the output of a language classification algorithm together with geolocation data. No prior work has considered using the web browser's \texttt{Accept-Language} as a feature of language classification.
	 
	 To the best of our knowledge, there exists a gap in the literature that we want to fill. This paper is first to present the performance of a keyword search using a hybrid method with strings tokenised into words and character based \ngram s and to present a potential improvement to language prediction in short messages by including country and \texttt{Accept-Language} header as predictor variables.

	\section{Greedy Tokenisation Algorithm}
	\label{sec:greedy}
	
	In our prior work \cite{8820497} we described our approach to a ticket classification system based on fuzzy string matching. A keyword search was performed by scanning a corpus of text in windows of the keyword's length. The string similarity was estimated using Cosine similarity, where both text and keyword were divided in \ngram \ s of size 2 (characters). Our prior work \cite{8820497} found that the Cosine algorithm was not only the most accurate but significantly faster (for two strings $n$ and $m$, the computational difficulty of the Cosine algorithm is $O(n + m)$ whilst the Edit Distance is $O(n \times n)$).
	The method was not sensitive to beginning and end of the string. In this Section we would like to present our solution to this issue.
	
	
	\subsection{Definition}
	\begin{algorithm}[h]
		
		\textbf{INPUT:} searched string $X$, text $Y$, similarity threshold $\theta$\\
		\textbf{OUTPUT:}  a similarity $S$ of the first match
		\begin{algorithmic}
			\State $l_X \gets$ length of X
			\State $l_Y \gets $ length of Y
			\State $c_X \gets$ word count of X
			\State $p_X \gets$ profile of X 
			\State $p_Y \gets$ profile of Y

			\For{$i \gets$ word from $P_Y$}
				\State $P_{Y_{-1}} \gets c_X - 1$ consecutive words in $P_Y$, starting with i
				\State $P_{Y_{0}} \gets c_X$ consecutive words in $P_Y$, starting with i
				\State $P_{Y_{+1}} \gets c_X + 1$ consecutive words in $P_Y$, starting with i
				\State $w_{-1} \gets$ Cosine similarity of $P_X$ and $P_{Y_{-1}}$  \Comment{*}
				\State $w_{0} \gets$ Cosine similarity of $P_X$ and $P_{Y_{0}}$  
				\State $w_{+1}\gets$ Cosine similarity of $P_X$ and $P_{Y_{+1}}$ 
				\State $S\gets max\{w_{-1}, w_{0}, w_{+1}\}$ 
				
				\If{$S \geq \theta$}
				\State return $S$
				\EndIf

			\EndFor
		\end{algorithmic}
		\caption{The Greedy algorithm for a keyword search.  
			* The similarity function runs only if the number of characters in $P_{Y_j}$ is within bounds $(1-\theta)* c_X \leq c_{P_{Y_{j}}} \leq (1+\theta)* c_X$, with $j = -1, 0, 1$. }
			\label{fig:dplcs}
	\end{algorithm}
	
	Our novel approach to fuzzy string matching consists of two tokenisation steps. In the first part we divide both strings into words by white spaces. We create a profile for both strings, which stores all words in the right order with the information of their lengths. In order to match on keywords which are divided into multiple words (e.g. \textit{nameservers} and \textit{name-servers}) we calculate the similarity for three cases:
	\begin{itemize}
	\item the searched string and a part of the scanned string one word shorter than the searched string,
	\item the searched string and a part of the scanned string of the same length as the searched string,
	\item the searched string and a part of the scanned string one word longer than the searched string.
	\end{itemize} 
We calculate a similarity only if the number of characters of the part of the scanned string is within the acceptable bounds wrt. to the similarity threshold $\theta$. We scan the ticket body ($Y$) word by word. In every iteration we choose the highest similarity $S$ from the three above-mentioned cases. If the condition $S \geq \theta$ if fulfilled, the scan stops and the function returns the similarity value. 

\subsection{Experiments}

	In order to measure accuracy of our new approach we gathered 1790 tickets falling into a chosen product category (known as "DNS"). The tickets were processed using the multi-classifier outlined in \cite{8820497} both before and after modification with our novel approach.

	With our novel approach; of the 1790 tickets, 1286 were properly classified as DNS, 249 were left unclassified, and 255 were misclassified (with majority being in three other product areas; Crypto (99), Server Errors (81) and Registrar (25)). The mean processing time was 0.012 seconds per ticket and median 0.00934 seconds per ticket.
	With our old method 1324 tickets we classified as DNS, 236 were left unclassified and 230 were misclassified. The mean processing time of the old approach was 0.073 seconds per ticket and median 0.056 seconds per ticket.
	As the new approach is sensitive to the beginning and the end of the string we miss a little on the coverage yet the improvement in computing time is valuable (the new approach is more than 6 times faster). 
	In order to estimate precision and recall we gathered 1208 tickets from a product category known as "Crypto". The same multi-classifier was applied. 
	
	\begin{table}[h]
		\caption{Precision and recall.}
		\label{tab:freq}
		\begin{tabular}{ccl}
			\toprule
			Method & Precision & Recall\\
			\midrule
			Old & 0.878 & 0.718 \\
			Greedy Approach & 0.855 & 0.740 \\
			\bottomrule
		\end{tabular}
	\end{table}
	
	\section{Metadata-Enhanced Language Classification}
	\label{sec:lang}
	
	So far, we have outlined an improved algorithm for fuzzy string matching, however improved language classification can be important to improving the accuracy of a given Natural Language Processing pipeline. In this instance, we take chat messages from real world customer support chatbot which is using the \cite{nlpjs} library for language classification. The classification is run on the first message sent by a user, given the need to understand if the user is using a supported language as early as possible in the conversation.
	
	We gathered $3204$ chats tickets alongside the language classification, the HTTP \texttt{Accept-Language} header presented by the users web browser and list the languages common to the visitor's country (using MaxMind GeoIP \cite{maxmindaddress} for geolocation based on their IP Address and open-source data \cite{countrydata} to get the languages associated with a given country).
	
	Predicting language from the first chat message, as we do here, can be challenging and will not replicate all use-cases. The first message could be just a simple greeting or could be a lengthy description of the issue the customer experiencing. As the chatbot is used for support purposes on a internet infrastructure product; in some cases visitors even included some software log lines on their first message on chat, making it harder to flag whether the visitor would like to get support in a non-English language.
	
\begin{figure}[h]
		\centering
		\includegraphics[width=1\linewidth]{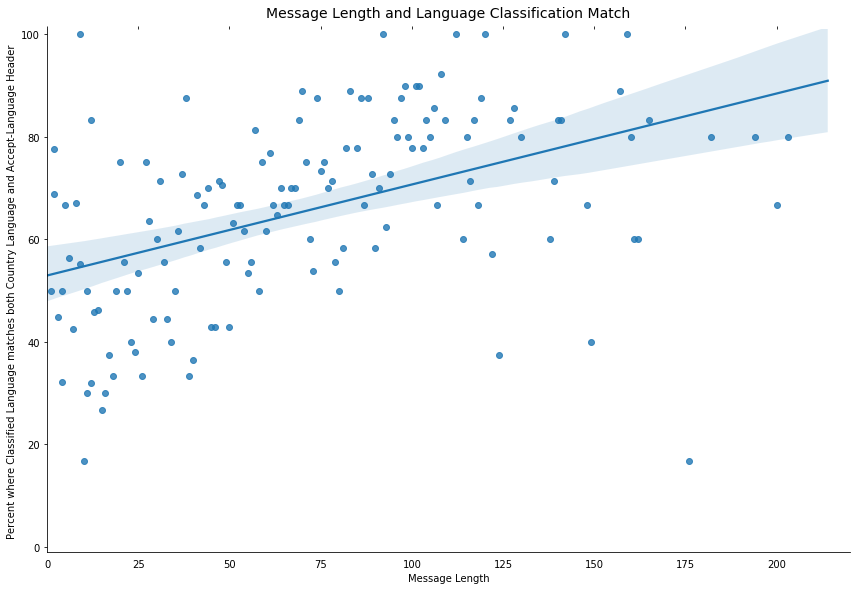}
		\caption[]{Message Length and Language Classification Match}
		\label{fig:sops-1786}
	\end{figure}

	In Fig \ref{fig:sops-1786}, we observe that as the length of a message increases in length - so does the probability that classified language will match the visitor's country language and their browsers \texttt{Accept-Language} header.
	
\begin{figure}[h]
		\centering
		\includegraphics[width=1\linewidth]{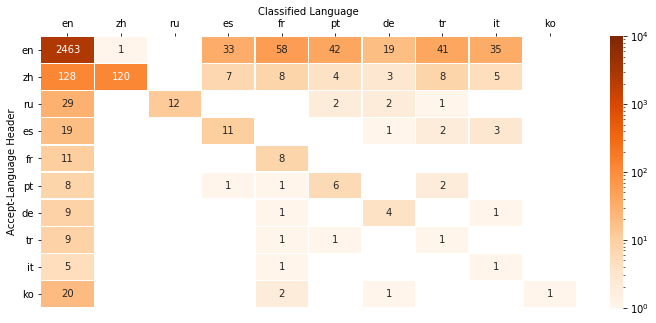}
		\caption[]{Classified Languages and Top 10 Accept-Language}
		\label{fig:sops-1780}
	\end{figure}

\begin{figure}[h]
		\centering
		\includegraphics[width=1\linewidth]{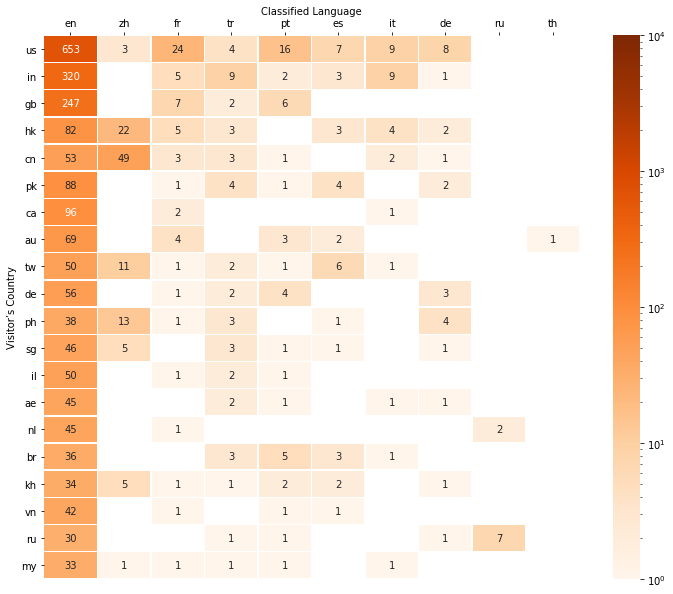}
		\caption[]{Classified Languages and Top 20 Countries}
		\label{fig:sops-1782}
	\end{figure}

	 We do also observe correlation between the classified language and \texttt{Accept-Language} header, as shown in Fig \ref{fig:sops-1780}. Although harder to visualise, a similar correlation can be observed between the classified language and the visitor's country in Fig \ref{fig:sops-1782}. In $67\%$ of chats, all parameters were in agreement. In $15\%$, the classified language only matched \texttt{Accept-Language} header and a further $5\%$ matched only the country languages. $13\%$ had no agreement between these three parameters. These results are visualised in Fig \ref{fig:sops-1782}.
	
	\begin{figure}[h]
		\centering
		\includegraphics[width=0.7\linewidth]{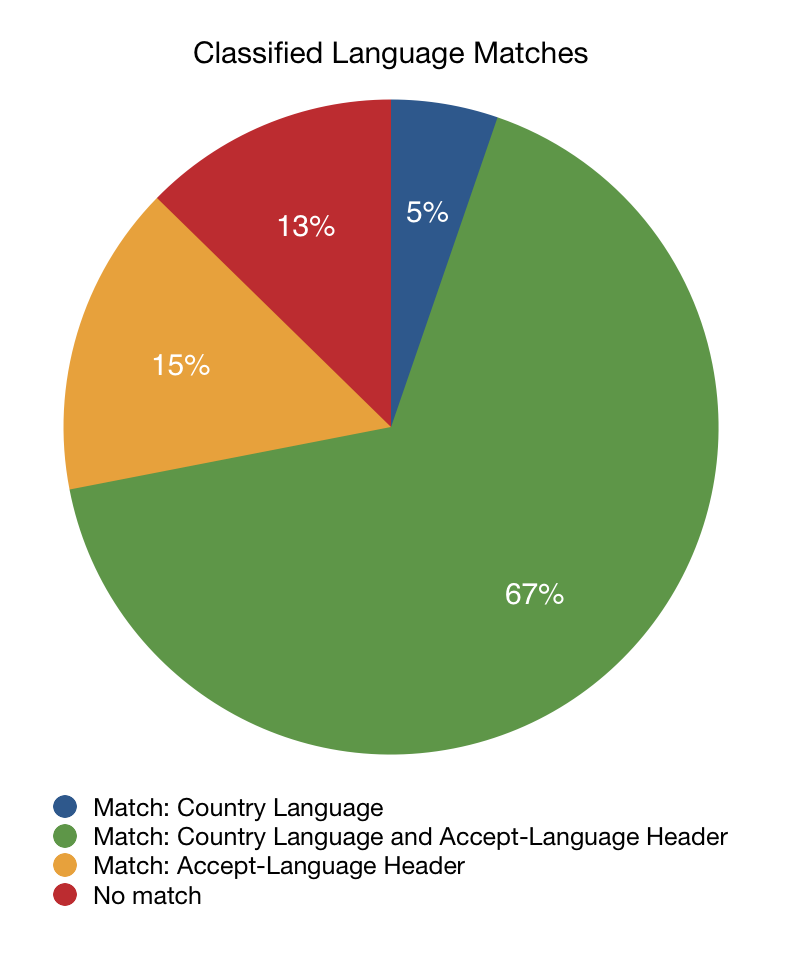}
		\caption[]{Classified Language Matches}
		\label{fig:sops-1785}
	\end{figure}

	The \texttt{Accept-Language} header was $14\%$ more likely to match the language classification than IP Address but $23\%$ more coverage was obtained by allowing either parameter to match the classified language than both, indicating both data sources can add value in a classification system. This observational evidence may be leveraged by future work to experiment different approaches to incorporate such metadata into novel language classification approaches.
	
	\section{Conclusions}
	\label{sec:conclu}
	In this paper, we presented a novel greedy tokenisation approach of strings for use in fuzzy keyword search. Our approach allows for efficient search of keywords using \ngram \ string similarity algorithms, even where keywords are subdivided into multiple words in the corpus text. Experimental results show that greedy tokenisation decreased processing time by $83.6\%$, we estimated an improvement in recall of $3.1\%$ with a decrease in precision of $2.6\%$.
	
	Further; we provide real-world observational data comparing classified languages to user metadata. Consistent with other reports of low classification accuracy on short strings, observed that the likelihood of the classified language matching the metadata increases with the message length. Whilst the \texttt{Accept-Language} matched the classified language in $82\%$ of instances, the country languages (extracted from the user's IP Address) only matched in $72\%$ of instances. This indicates that \texttt{Accept-Language} likely provides a better signal than IP Address for user language, but whilst all three signals only matched in $67\%$ of instances, $87\%$ coverage can be obtained if the classified language is allowed to match either \texttt{Accept-Language} header or the country language.
	
	Whilst further research is needed; our data supports future research into using metadata to support language classification algorithms, particularly for short messages or instances where higher certainty is needed before making language classification decisions. One potential area of study is the creation of a model that receives input from language classification algorithms as well as different sources of metadata, with message length potentially being a further dimension.

	\bibliographystyle{unsrt}
	\bibliography{bibfile}
	
\end{document}